\pdfoutput=1

\documentclass[11pt]{article}

\usepackage[final]{acl}

\usepackage{times}
\usepackage{latexsym}
\usepackage{times}
\usepackage{latexsym}
\usepackage{array}
\usepackage{graphicx}
\usepackage{xcolor}
\usepackage{caption} 

\usepackage{enumitem}
\usepackage{hyperref}

\usepackage[most]{tcolorbox}
\usepackage{mathtools}
\usepackage{enumitem}
\usepackage{colortbl}
\usepackage{diagbox}
\usepackage{wrapfig}
\usepackage{pst-node} 
\usepackage[T1]{fontenc}
\usepackage{amsmath}

\usepackage[utf8]{inputenc}

\usepackage{microtype}

\usepackage{inconsolata}

\usepackage{tabularx}
\usepackage{multirow}
\usepackage{booktabs}
\usepackage[normalem]{ulem}
\usepackage{xspace}
\renewcommand{\footnotesize}{\small}
\renewcommand{\arraystretch}{0.9}

\usepackage{url}

\usepackage[T1]{fontenc}

\usepackage[utf8]{inputenc}

\usepackage{microtype}

\usepackage{inconsolata}

\usepackage{graphicx}
\usepackage{textcomp} 
\title{Hybrid Graphs for Table-and-Text based Question Answering using LLMs}

\usepackage{authblk} 

\author[1]{\textbf{Ankush Agarwal} \thanks{Corresponding author : \texttt{ankush.agarwal@fujitsu.com}}}
\author[1,2]{\textbf{Ganesh S}\thanks{Work done during internship at Fujitsu Research India.}}
\author[1]{\textbf{Chaitanya Devaguptapu}}

\affil[1]{Fujitsu Research India}
\affil[2] {IIT Madras}
\affil[ ]{\texttt{ankush.agarwal@fujitsu.com, ganeshsenrayan@outlook.com, email@chaitanya.one}}

\newcommand{\name}{ODYSSEY\xspace}

\begin{document}
\maketitle
\begin{abstract}

Answering questions that require reasoning and aggregation across both structured (tables) and unstructured (raw text) data sources presents significant challenges. Current methods rely on fine-tuning and high-quality, human-curated data, which is difficult to obtain. Recent advances in Large Language Models (LLMs) have shown promising results for multi-hop question answering (QA) over single-source text data in a zero-shot setting, yet exploration into multi-source Table-Text QA remains limited.
In this paper, we present a novel Hybrid Graph-based approach for Table-Text QA that leverages LLMs without fine-tuning. Our method constructs a unified Hybrid Graph from textual and tabular data, pruning information based on the input question to provide the LLM with relevant context concisely.
We evaluate our approach on the challenging Hybrid-QA and OTT-QA datasets using state-of-the-art LLMs, including GPT-3.5, GPT-4, and LLaMA-3. Our method achieves the best zero-shot performance on both datasets, improving Exact Match scores by up to 10\% on Hybrid-QA and 5.4\% on OTT-QA. Moreover, our approach reduces token usage by up to 53\% compared to the original context.

\end{abstract}

\section{Introduction}
\begin{figure}[h]
    \includegraphics[width=\linewidth]{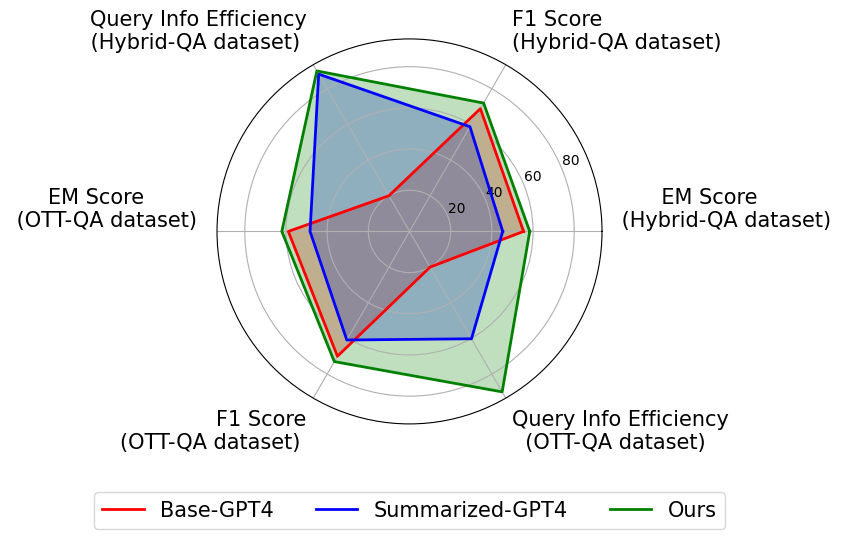}
    \caption{\textbf{ Multi-Dimensional Improvements}: Our method (with GPT-4 as reader LLM) demonstrates superior results on Hybrid-QA and OTT-QA. \textbf{Metrics used:} EM: Exact-Match with the gold answer, F1-Score, Query Info Efficiency: normalized metric ($\frac{1}{\text{Input Token Size}}$) that quantifies the efficiency of using fewer input tokens to represent the same documents, w.r.t. reader LLM.}
  \label{fig:spider_plot}
\end{figure}
In today's data-rich world, information is scattered across various sources, which can be broadly divided into two types: structured (tables, databases) and unstructured (raw text from various files). The ability to effectively answer questions that require reasoning and aggregation across diverse data sources has become increasingly crucial. Such questions are often referred to as \textit{hybrid questions} and the task is often referred to as Table-Text Question Answering (QA). For instance, consider the question: \textit{"What place was achieved by the person who finished the Berlin marathon in 2:13.32 in 2011, the first time he competed in a marathon?"}, answering this question requires extracting the name of the person who finished the Berlin marathon in 2011 from raw unstructured text and linking it with a structured data source like a database to determine their place.  The ability to answer hybrid questions has immense real-world value, as it enables the combination of relevant information from multiple sources, leading to more comprehensive and capable QA systems.

Existing QA methods primarily focus on single data sources, either structured or unstructured.  These approaches are limited when answering questions that require reasoning and aggregation across \textit{\textbf{both}} structured and unstructured data. Datasets like Hybrid-QA~\cite{chen2020hybridqa} and OTT-QA~\cite{chen2020open} necessitate combining information from tables and text, however current methods applied to these datasets~\cite{lee2023mafid, eisenschlos2021mate, kumar2023multi, lei2023s3hqa} rely on the availability of training data with pre-established connections between the structured and unstructured data. In real-world applications, such training data may not always be available and is often expensive to collect. For example, even for general data sources like Wikipedia, there are only two publicly available datasets (Hybrid-QA and OTT-QA) that contain hybrid questions.

Given sufficient context, large language models (LLMs) ~\cite{team2023gemini, achiam2023gpt} can effectively extract answers from the provided context for a wide range of question categories and domains. Recently, LLM-based zero-shot QA has emerged as a robust alternative to traditional fine-tuning-based QA methods. For the task of Table-Text QA, a straightforward approach is to provide the entire text and table as input to the LLM alongside the question. However, this method is not a scalable approach due to token cost and the limited context length of LLMs. Approaches such as context truncation ~\cite{xie2022unifiedskg} and summarization ~\cite{jin2024comprehensive} using off-the-shelf techniques like those in LangChain\footnote{\url{https://www.langchain.com}}  have been proposed. Nevertheless, indiscriminate truncation or summarization can lead to significant performance degradation, as critical information may be omitted, resulting in less accurate or incomplete answers (as shown in  Section-\ref{sec:results}, Table-\ref{tab:comparison}).

\begin{figure*}
  \centering
  \footnotesize
    \includegraphics[width=0.88\textwidth]{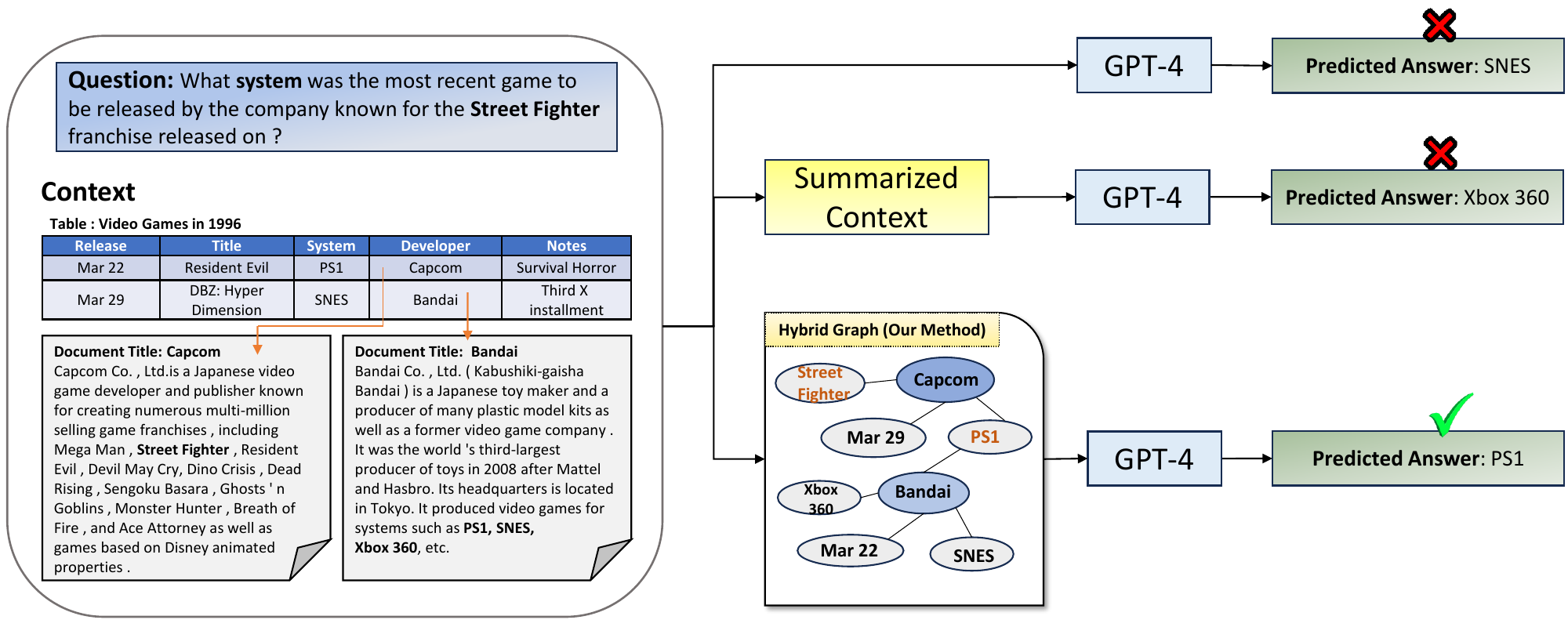} 
\caption{\textbf{Case study on Hybrid-QA:} Comparison of our method (\name) against various baselines on an example from the Hybrid-QA dataset. Baselines: (i) \textbf{Question + Context:} Providing the LLM only the question without any additional context (ii) \textbf{Question + Summarized Context:} Passing the question along with the summarized documents and table. 
Our method delivers accurate answer because the Hybrid Graph efficiently connects "Street Fighter" from the document "Capcom" with the relevant table, guiding GPT-4 in generating the correct response, i.e., "PS1" from the "System" table column.
}
  \label{fig:case-study}
\end{figure*}

To enhance the ability of LLMs to answer hybrid questions and efficiently retrieve and leverage relevant context, we propose \name \footnote{Named after the primary goal of our approach to navigate from complex multi-hop Table-Text data to hybrid graphs that provide greater clarity}.  Our approach operates in a zero-shot setting, aiming to filter out noise and provide the LLM with the most relevant context in a concise manner. At a high level, our method consists of two main steps: i) Constructing a \textit{\textbf{single}} Hybrid Graph from both textual and tabular data, and ii) Pruning information from the graph based on the question. To demonstrate the effectiveness of our method compared to existing approaches, we present a case study on the Hybrid-QA dataset \cite{chen2020hybridqa} in Figure \ref{fig:case-study}.

We evaluate \name on Hybrid-QA~\citet{chen2020hybridqa} and OTT-QA~\citet{chen2020open}  datasets using three state-of-the-art LLMs for QA: GPT-3.5, GPT-4 and LLaMA-3. We compare our method against relevant baselines across five metrics: Exact Match (EM) \cite{zhang2023tablellama}, F1-Score \cite{lei2023s3hqa}, Precision, Recall, and BERTScore-F1 \cite{zhangbertscore}, demonstrating that constructing and leveraging a Hybrid-Graph by combining structured and unstructured information sources results in significant performance gains. Our key contributions can be summarized as follows:

\begin{itemize}[nosep]
    \item A novel approach that jointly distills information from structured and unstructured data sources to construct a Hybrid Graph.
    \item An increase in performance over the current SoTA fine-tuning-free approach, improving EM and F1 scores by 7.3\% and 20.9\% for Hybrid-QA using GPT-4 (see Table \ref{tab:hybridqa_methods}).
    \item A significant reduction in the input token size. Our Hybrid Graph based approach uses up to 45\% and 53\% fewer tokens than the original table and text for the Hybrid-QA and OTT-QA datasets respectively (see Table \ref{tab:total_token_count}).
\end{itemize}






\section{Related Work}
\label{sec: Related_Work}

\textbf{Question Answering:} With the advent of language models, the task of question answering over structured and unstructured data sources has gained a significant traction and interest. 
For structured data sources like tables, recent methods have focused on generating SQL queries from natural language input ~\cite{wang2020rat, hui2022s, li2023resdsql, li2023graphix, gao2023text}, converting structured data to graphical forms \cite{jiang2023structgpt, perozzi2024let, tan2024struct}, or allowing language models to directly interact with the tables ~\cite{herzig2020tapas, wang2021tuta}. 
Similarly, efforts have been made to develop efficient QA systems for unstructured text ~\cite{seo2016bidirectional, hu2018reinforced, perez2020unsupervised, seonwoo2020context}.
While many approaches have been proposed for answering questions on either structured or unstructured data sources while treating them in isolation, there is a scarcity of methods that focus on Table-Text QA systems, \textit{i.e.}, questions that require reasoning and aggregation of information from both structured and unstructured data sources simultaneously.


\noindent \textbf{Table-Text Question Answering:} The task of Table-Text QA based on text-and-tabular data has recently started gaining traction because of the creation and availability of datasets like Hybrid-QA \cite{chen2020hybridqa} and OTT-QA \cite{chen2020open}, both of which are based on Wikipedia as the primary data source containing both text and tabular data. 


\noindent To tackle challenges associated with Table-Text QA, several approaches have been proposed \cite{chen2020hybridqa, wang2022muger2, lei2023s3hqa}. While these methods perform well, they rely on supervised fine-tuning data to learn and aggregate information across text and tables for question answering. Collecting supervised data for the task of Table-Text QA is not only resource intensive~\cite{orr2023social} but may not always be possible. 
\\
\noindent \textbf{Fine-Tuning-free Table-Text QA:} The emergence of large-scaled pre-trained LLMs~\cite{achiam2023gpt, le2023bloom} has revolutionized the field of NLP and especially has opened up ways to address the task of QA in a fine-tuning-free manner. Based on a given question, the relevant context is retrieved from the data corpus using state-of-the-art retrievers~\cite{robertson2009probabilistic, karpukhin2020dense, khattab2020colbert} and this context is passed along with the question to the LLM. Recent work \cite{shi2024exploring} tackles this issue by generating and executing code in a few-shot setting, similar to natural language-to-SQL conversion, but it does not account for noise in the context.
\\
\name is a fine-tuning-free zero-shot approach for Table-Text QA. Our method efficiently prunes relevant information from both tabular and textual sources and performs the Table-Text QA task. Furthermore, \name effectively addresses the unique challenges associated with Table-Text QA, such as integrating information from heterogeneous sources and performing multi-hop reasoning across tables and text.

\section{Methodology}

\begin{figure*}
  \centering
  \footnotesize
    \includegraphics[width=0.88\textwidth]{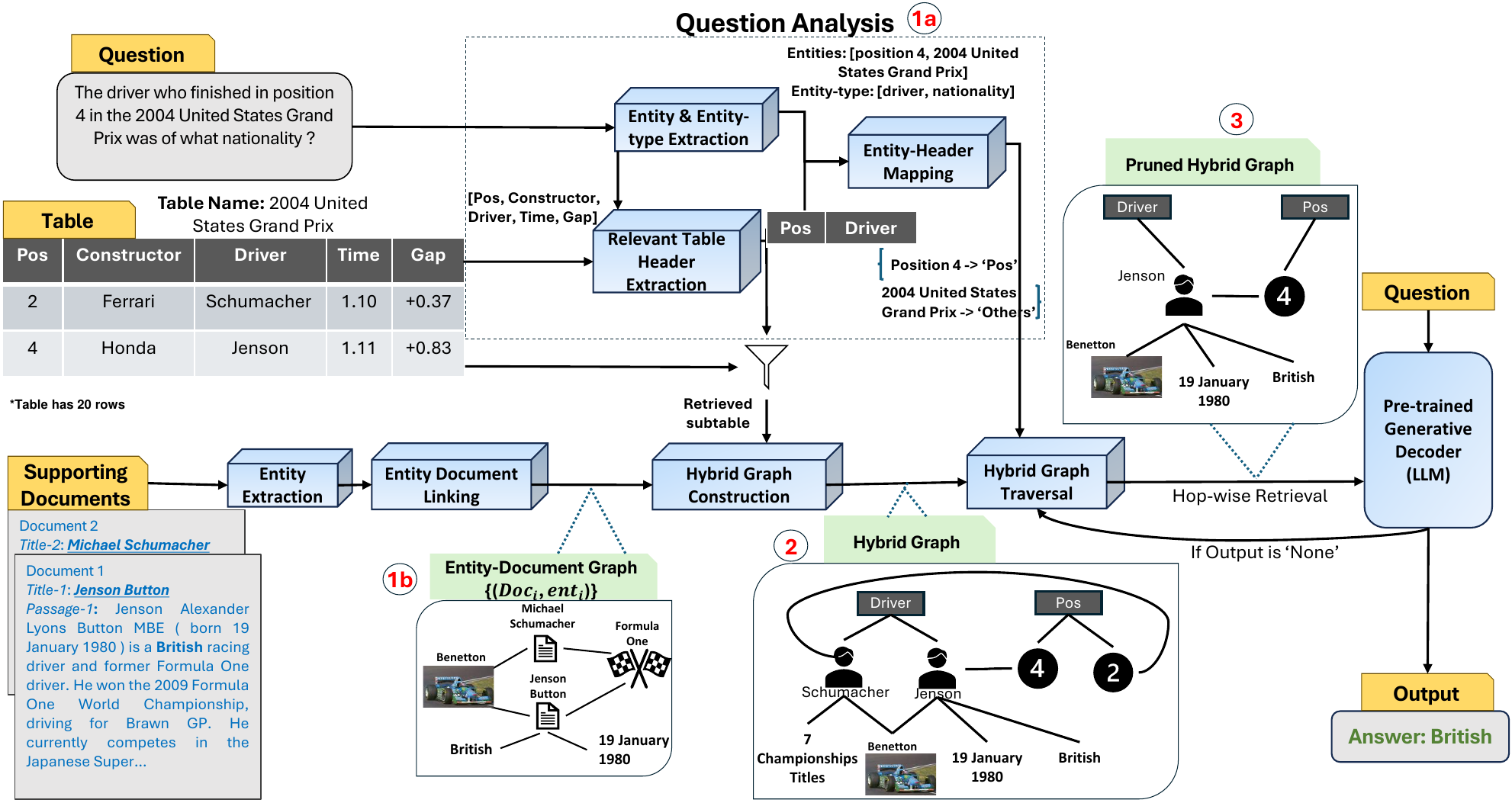} 
\caption{\textbf{Overview of the \name framework.} 
Our method comprises of 3 steps: i) Question Analysis, ii) Hybrid Graph Construction, and iii) Hybrid Graph Traversal. First, we begin with Question Analysis ({\large{$\textcircled{\small{\textcolor{red}{\textbf{1a}}}}$}} in the figure) from where we get question entities, retrieved sub-table, and entity-header mapping. Next, we construct the Entity-Document Graph ({\large{$\textcircled{\small{\textcolor{red}{\textbf{1b}}}}$}} in the figure). Using entity-doc graph and retrieved sub-table, we construct the Hybrid Graph ({\large{$\textcircled{\small{\textcolor{red}{\textbf{2}}}}$}} in the figure). At last, we perform Hybrid Graph Traversal ({\large{$\textcircled{\small{\textcolor{red}{\textbf{3}}}}$}} in the figure) to get the pruned graph which serves as input for the LLM. For a detailed walkthrough, refer to Appendix \ref{sec: walkthrough}}
  \label{fig:approach}
\end{figure*}

In this section, we explain the key parts of our approach for solving Hybrid-QA in a fine-tuning-free manner. We begin by describing the problem setting, followed by a detailed explanation of our method, which includes question analysis, graph construction and graph traversal. Finally, we discuss the retrieval and the information that is passed into the LLM.


\subsection{Problem Definition} \label{sec: ProblemD}
Assume a structured data source $\mathcal{T}$ with rows denoted using $\mathcal{R}$, where $r_i \in \mathcal{R}$ denotes the $i^{\text{th}}$ row, and headers denoted using $\mathcal{H}$, where $h_j \in \mathcal{H}$ denotes the $j^{\text{th}}$ header. Let $\mathcal{T}(r_i, h_j)$ represent the value in the $i^{\text{th}}$ row and the $j^{\text{th}}$ header / column. For a few cells in every row, certain cells in column $h_j$ contain unstructured data sources linked to them. Specifically, $\mathcal{T}(r_i, h_j)$ is linked to $k$ documents $\{d_{ij}^1, d_{ij}^2, \dots, d_{ij}^k\}$ which contain unstructured information.

Given a natural language question $\mathcal{Q}$ related to both the structured data $\mathcal{T}$ and the unstructured data $\mathcal{D}$ (comprising the linked documents), the objective is to output an answer $\mathcal{A}$.

To aid in the explanation of our methodology, we will use the following example throughout this section.
Consider the following question $\mathcal{Q}$ from one of the datasets we consider in our experiments: \textit{"The driver who finished in position 4 in the 2004 United States Grand Prix was of what nationality?"}. This question requires reasoning and aggregation across a structured data source and an unstructured data source. The structured data source, table $\mathcal{T}$, contains the following headers $\mathcal{H}$ = \{\texttt{`Pos', `Constructor', `Driver', `Time', `Gap'}\} with $20$ rows ($|\mathcal{R}| = 20$), and each row $r_i \in \mathcal{R}$ contains 5 cells $\mathcal{T}(r_i, h_j)$ for $j = 1, \dots, 5$. Documents linked to a cell $\mathcal{T}(r_i, h_j)$ in the table are denoted as $\mathcal{D}(r_i, h_j) = \{d_{ij}^1, d_{ij}^2, \dots, d_{ij}^k\}$. Our objective is to generate the answer $\mathcal{A}$ using the language model $\mathcal{L}$, which spans table cells and linked documents.

\subsection{Proposed Method (\name)}
\label{sec: proposed_method}

Our proposed method consists of the following key steps:
\begin{enumerate}[nosep]
\item \textbf{Question Analysis:} We analyze the question to identify key entities and entity types that will later aid in graph construction, traversal, and answering the question (Section \ref{QAnalysis}).
\item \textbf{Hybrid Graph Construction:} We locate the tabular evidence through structure modeling and model the relationships among heterogeneous evidence (Section \ref{HGraph}).
\item \textbf{Hybrid Graph Traversal:} We prune the Hybrid Graph based on question-derived entities and perform multi-hop reasoning and traversal (Section \ref{HTraversal}).
\item \textbf{Reader LLM:} We use a Language Model to generate the final answer using the pruned graph and linked documents (Section \ref{LlmReader}).
\end{enumerate}

\subsubsection{Question Analysis} \label{QAnalysis}

To effectively address questions based on a given context, our approach utilizes an LLM to identify entities and entity types within the question. 
For example, in figure \ref{fig:approach} - {\large{$\textcircled{\small{\textcolor{red}{\textbf{1a}}}}$}}, we identify entities such as \texttt{`position 4'} and \texttt{`2004 United States Grand Prix'}, along with entity types \texttt{`driver'} and \texttt{`nationality'}.
These entities and entity types guide us in determining relevant table headers, such as \texttt{`Pos'} and \texttt{`Driver'}, where \texttt{`Pos'} corresponds to \texttt{`position 4'} and \texttt{`Driver'} corresponds to the entity type.
We establish an entity-header mapping by aligning entities with the retrieved relevant headers. Entities that match specific headers are mapped accordingly, while those that do not fit any header are categorized as  \texttt{`Others'} (see figure \ref{fig:approach} - {\large{$\textcircled{\small{\textcolor{red}{\textbf{1a}}}}$}}). The effectiveness of this mapping technique will be demonstrated in Section \ref{HTraversal}.
The Question Analysis prompt used in our method is detailed in Appendix \ref{sec: QAnalysis_prompt}.

\subsubsection{Hybrid Graph Construction} \label{HGraph}

Our Hybrid Graph consists of two connected components: the table and documents. By selecting relevant table headers, we retrieve the sub-table. Simultaneously, from the documents, we construct the Entity-Document graph. These components are then integrated to form the Hybrid Graph.

\noindent \textbf{Sub-table Retrieval}  
Column headers are selected to retrieve the sub-table. As explained in Section \ref{QAnalysis}, the \texttt{`Pos'} and \texttt{`Driver'} table header columns are selected for the Hybrid Graph. Each row cell in the table is connected with others, and we replicate these connections to visualize the sub-table in a graph format. Refer to Figure \ref{fig:approach} - {\large{$\textcircled{\small{\textcolor{red}{\textbf{2}}}}$}} to see the connections between the row cells.

\noindent \textbf{Entity-Document Graph Construction}
The entity-document graph includes document and entity nodes, connected by single edges. Named entities are extracted from documents, and edges are formed between document and entity nodes, creating a bipartite graph that shows their relationships (see figure \ref{fig:approach} - {\large{$\textcircled{\small{\textcolor{red}{\textbf{1b}}}}$}}).
\\
\noindent As discussed in Section \ref{sec: ProblemD}, certain table cells are linked to specific documents. We connect the entities from each document to their corresponding retrieved sub-table cells to form the Hybrid Graph (see figure \ref{fig:approach} - {\large{$\textcircled{\small{\textcolor{red}{\textbf{2}}}}$}}).

\subsubsection{Hybrid Graph Traversal} \label{HTraversal}
After constructing the Hybrid Graph, we prune it based on the question to filter out noise. Using the entity-header mapping dictionary (described in Section \ref{QAnalysis}), we perform a Breadth-First Search (BFS) to semantically match the question entities with table column cells and entities in the entity-document graph.
For semantic matching, we first gather all document entities into a set called \texttt{entity\_total}. In the entity-header mapping, entities mapped to a header are aligned with the corresponding column, while those mapped to \texttt{`Others'} are aligned with \texttt{entity\_total} (as explained in Section \ref{QAnalysis}). For example, as shown in Figure \ref{fig:approach}, the entity \texttt{`Positon 4'} is mapped to the header \texttt{`Pos'}, and after semantic matching, it aligns with the entity `4'. In contrast, the entity \texttt{`2004 United States Grand Prix'} does not match any entity in \texttt{entity\_total}.
If the header-mapped entities are not successfully matched, we expand the matching process to include other table columns to account for potential noise in the entity-header mapping.
\\
Once the semantically matched entities are obtained, we initiate a BFS traversal using them as starting points. In figure \ref{fig:approach}, the starting point is entity \texttt{`4'}. 
We perform a 3-hop traversal, which results in a pruned graph (as shown in Figure \ref{fig:approach} - {\large{$\textcircled{\small{\textcolor{red}{\textbf{3}}}}$}}). During traversal, we record the paths in the graph and store them in a hop-wise dictionary, categorized by 1-hop, 2-hop, and 3-hop.
If no entities are matched during semantic matching, we store the entire table and linked passages.  The number of such cases is mentioned in Section \ref{table-text_results}.

\subsubsection{Reader LLM prompt} \label{LlmReader}


After obtaining the pruned graph and storing it in a hop-wise dictionary, we use an LLM to answer the question. Initially, only 1st-hop items are provided as context. If no answer is found, the LLM returns \texttt{`None'}. 
We then include 2nd-hop items as context, along with relevant linked passages given as output from the 1st-hop LLM call, and concatenate the 1st-hop and 2nd-hop tables. This iterative process continues up to 3 hops. If the LLM still returns \texttt{`None'} after 3 hops, the entire table-text is used as context. The number of such cases is mentioned in Section \ref{table-text_results}.
We limit the process to 3 hops due to minimal improvements observed in LLM performance beyond that during hyperparameter tuning (see Section \ref{sec: exp_setup}). The LLM reader prompt used in our method is detailed in Appendix \ref{sec:prompt_reader}.


\section{Experimental Setup}
\label{sec: exp_setup}

This section details our experimental setup, including the datasets used to evaluate our method, the baselines for comparison, evaluation metrics, and implementation specifics.

\subsection{Experimental Datasets}
\label{sec: dataset}
We evaluate \name on two Table-Text Hybrid-QA datasets: Hybrid-QA \cite{chen2020hybridqa} and OTT-QA \cite{chen2020open}.
A detailed explanation of the experimental datasets is provided in Appendix \ref{sec: dataset}.


\noindent \textbf{Hybrid-QA} is a large-scale, complex, multihop Table-Text Hybrid-QA benchmark comprising tables and texts from Wikipedia. Each table row describes various attributes of an instance, linked to corresponding Wikipedia passages that provide detailed descriptions. Dataset statistics are provided in Table \ref{tab: hybrid_stats}.

\begin{table}[h!]
\centering
\begin{tabular}{lrrrr}
\toprule
Split & Train & Dev & Test & Total \\
\midrule
In-Passage & 35,215 & 2,025 & 20,45 & 39,285  \\
In-Table & 26,803 & 1,349 & 1,346 & 29,498 \\
Computed & 664 & 92 & 72 & 828 \\
\midrule
Total & 62,682 & 3,466 & 3,463 & 69,611 \\
\bottomrule
\end{tabular}
\caption{Data Split: In-Table means the answer comes from plain text in the table, and In-Passage means the answer comes from certain passage.}
\label{tab: hybrid_stats}
\end{table}

\noindent \textbf{OTT-QA} extends Hybrid-QA into a large-scale open-domain QA dataset over tables and text, requiring both table and passage to be retrieved before answering questions. It samples around 2,000 questions from the in-domain Hybrid-QA dataset, mixing them with newly collected out-domain questions for the dev and test sets. OTT-QA consists of 41,469 questions in the training set, 2,214 in the dev set, and 2,158 in the test set. Unlike Hybrid-QA, many questions in OTT-QA have multiple plausible inference chains. Dataset statistics are provided in Table \ref{tab: ott_stats}.

\begin{table}[h!]
\centering
\begin{tabular}{lrrr}
\toprule
 & Train & Dev & Test \\
\midrule
Total & 41,469 & 2,214 & 2,158 \\
\bottomrule
\end{tabular}
\caption{OTT-QA Statistics}
\label{tab: ott_stats}
\end{table}

 Since we operate in a zero-shot setting, we do not use the training set. Instead, we uniformly sample 500 examples from the dev sets of both Hybrid-QA and OTT-QA for comparison with baselines. For hyperparameter tuning, we select 50 questions from the Hybrid-QA dev set, ensuring they are distinct from our experimental set. For OTT-QA, we choose the newly added questions different from Hybrid-QA and use the retrieved table and retrieved text \cite{chen2020open} to test our methodology. The decision to use a sample dataset \cite{trivedi2023interleaving, li2023leveraging} for testing our methodology is driven   
by the costs associated with OpenAI API\footnote{\url{https://openai.com/api/}}.

\subsection{Baselines} \label{sec:baselines}
 We operate in a fine-tuning-free, zero-shot and closed-domain setting. Our experiments involve both closed-source and open-source large language models. For closed-source models we utilize two popular LLMs - GPT-3.5-turbo-1106\footnote{\url{https://platform.openai.com/docs/models/gpt-3-5-turbo}} and GPT-4-1106-preview~\cite{achiam2023gpt}, both at a temperature setting of $0$. For open source LLMs, we employ Llama3-8B\footnote{\url{https://huggingface.co/meta-llama/Meta-Llama-3-8B}}. We compare our method against three baselines:
\\
\begin{itemize}[nosep]
    \item \textbf{Base}: To test the parametric knowledge of the reader LLM, we provide only the question and elicit a response.
    \item \textbf{Base w/ Table \& Text} \cite{zhang2023tablellama}: To test the hybrid tabular and textual QA understanding of the LLM, we provide the full context (table and passages) following the prompt outlined in ~\citet{zhang2023tablellama}.
    \item \textbf{Base w/ Table \& summarized Text}\footnote{\url{https://blog.langchain.dev/semi-structured-multi-modal-rag/}}: To reduce the noise in the unstructured information and to make it easier for LLM to find the answer, we employ LangChain's proposed approach to summarize \cite{jin2024comprehensive} the text and then pass the summary of the text with the entire table to the LLM for answering the questions.
    \item We also show results when both the text and the table are summarized. Refer to Appendix \ref{sec: summarizing} for details.
\end{itemize}
\noindent All prompts used for the above mentioned baselines are detailed in Appendix \ref{sec:baseline_prompts}. 

\subsection{Evaluation Metrics} \label{sec:eval-metrics}

For evaluation, we employ several metrics: Exact Match (EM) \cite{zhang2023tablellama}, F1-Score \cite{lei2023s3hqa}, Precision (P), and Recall (R) to assess the correctness of predicted answers. These metrics are implemented using the same codebase as Hybrid-QA \cite{chen2020hybridqa}. 
Additionally, for semantic evaluation, we employ BERTScore-F1 (B) \cite{zhangbertscore} and utilize the \texttt{bert-base-uncased} model \cite{devlin2018Bert} for computing similarity matching.

\subsection{Implementation Details}
\label{sec: imp_details}

\noindent \textbf{\name hyperparameters} 
We use GPT-3.5-turbo-1106 for question analysis, including entity extraction, relevant header fetching from tables, and entity-header mapping. To fetch entities from linked passages, we use the SpaCy\footnotemark[7]\footnotetext[7]{\url{https://pypi.org/project/spacy-transformers/}} transformers model with \texttt{en\_core\_web\_trf}\footnotemark[8]\footnotetext[8]{\url{https://huggingface.co/spacy/en_core_web_trf}}, which utilizes a RoBERTa-base \cite{liu2019roberta} model. These components collectively enable us to construct the Hybrid Graph.

For graph traversal, we match question entities with the Hybrid Graph using the \texttt{instructor-xl} model \footnotemark[9]\footnotetext[9]{\url{https://huggingface.co/hkunlp/instructor-xl}},  running on a GPU with 6GB of VRAM. We select the highest-ranked entity surpassing a 0.8 threshold. During the breadth-first search traversal, we store up to 3-hops to pass to the LLM reader. These values were determined by experimenting on 50 samples from the dev set.
The average runtime of \name is 3.6 seconds per instance.

\textbf{Text Summarization} To implement the Base with Table and Summarized Text, we use GPT-3.5-turbo-1106 for summarizing the text. Alternatively, this summarization can be done using fine-tuned models like T5 \cite{raffel2020exploring}.

\section{Results and Analysis} \label{sec:results}

In this section, we presents the results of our method and compare it with the baselines discussed in Section \ref{sec:baselines}. We show results using Llama3-8B, GPT-3.5 and GPT-4, along with the fine-tuned models on Hybrid-QA and OTT-QA datasets (see Table \ref{tab:comparison}, \ref{tab:hybridqa_methods}, \ref{tab:ottqa_methods}) across various evaluation metrics discussed in Section \ref{sec:eval-metrics}. 
Additionally, we analyze the token efficiency, \textit{i.e.}, the number of tokens passed as input to the LLM (see Table \ref{tab:total_token_count}). 
Furthermore, we provide a detailed analysis of our method, including an ablation study, error analysis (shown in Appendix \ref{sec: error}), and a hop-wise breakdown of the results, which are illustrated in Figure \ref{fig:hopwise_plot}.



\subsection{Evaluation on Tabular-and-Text QA}
\label{table-text_results}


\renewcommand{\arraystretch}{1.1} 
\begin{table*}[t]
\footnotesize
\captionsetup{font=footnotesize}
\centering
\begin{tabularx}{\textwidth}{l*{5}{>{\centering\arraybackslash}X}*{5}{>{\centering\arraybackslash}X}}
\cline{1-11}     
\multicolumn{1}{l}{\textbf{Datasets}} 
& \multicolumn{5}{c}{Hybrid-QA} 
& \multicolumn{5}{c}{OTT-QA} 
\\
\cmidrule(lr){2-6} \cmidrule(lr){7-11} 
\multicolumn{1}{l}{\textbf{ Methods}}
& \multicolumn{1}{p{0.8cm}}{\centering EM\\($\uparrow$)} 
& \multicolumn{1}{p{0.8cm}}{\centering F1\\($\uparrow$)} 
& \multicolumn{1}{p{0.8cm}}{\centering P\\($\uparrow$)}
& \multicolumn{1}{p{0.8cm}}{\centering R\\($\uparrow$)}
& \multicolumn{1}{p{0.8cm}}{\centering B\\($\uparrow$)}
& \multicolumn{1}{p{0.8cm}}{\centering EM\\($\uparrow$)} 
& \multicolumn{1}{p{0.8cm}}{\centering F1\\($\uparrow$)} 
& \multicolumn{1}{p{0.8cm}}{\centering P\\($\uparrow$)} 
& \multicolumn{1}{p{0.8cm}}{\centering R\\($\uparrow$)}
& \multicolumn{1}{p{0.8cm}}{\centering B\\($\uparrow$)}
\\ 
\cline{1-11}
\cline{1-11}
\rowcolor{gray!15} 
\multicolumn{11}{c}{\centering \fontfamily{lmss}\selectfont{ \textit{Reader: gpt-4-1106-preview (zero-shot)}}} \\
\cline{1-11}

Base & 4.60 & 12.44 & 12.08 & 12.25 & 62.10 & 4.85 & 12.44 & 12.25 & 14.24 & 64.30
\\
\arrayrulecolor{black!5}\hline

Base w/ Table \& Text & \multirow{2}{*}{55.40} & \multirow{2}{*}{68.84} & \multirow{2}{*}{68.92} & \multirow{2}{*}{71.79} & \multirow{2}{*}{85.54} &  \multirow{2}{*}{58.86} &
\multirow{2}{*}{72.28} &
\multirow{2}{*}{72.16} &
\multirow{2}{*}{74.28} & 
\multirow{2}{*}{87.51}
\\ \cite{zhang2023tablellama}
\\
\hline
\arrayrulecolor{black!5}\hline
Base w/ Table & \multirow{2}{*}{45.29} & \multirow{2}{*}{58.72} & \multirow{2}{*}{58.78} & \multirow{2}{*}{61.39} & \multirow{2}{*}{81.14} &
\multirow{2}{*}{48.31} &
\multirow{2}{*}{60.90} &
\multirow{2}{*}{61.47} &
\multirow{2}{*}{63.12} & 
\multirow{2}{*}{82.02}
\\ \& Summarized Text\setcounter{footnote}{1}\footnotemark 
\\
\arrayrulecolor{black}\hline
\rowcolor{blue!6}

Our Method w/o hopwise & 58.20 & 71.54 & 71.75 & 74.35 & 86.30 &  61.00 & 72.64 & 73.60 & 74.27 & 88.06
\\

\rowcolor{blue!6}
Our Method w/ hopwise 
& \textbf{58.40} & \textbf{71.80} & \textbf{71.62} & \textbf{74.22} & \textbf{86.53} & 
\textbf{62.02} & \textbf{73.02} & \textbf{73.40} & \textbf{75.13} & \textbf{88.18}
\\
\arrayrulecolor{black}\hline
\rowcolor{gray!15}
\multicolumn{11}{c}{\centering \fontfamily{lmss}\selectfont{ \textit{Reader: gpt-3.5-turbo-1106 (zero-shot)}}} \\
\cline{1-11}

Base & 4.20 & 11.54 & 11.62 & 12.45 & 65.05 & 5.27 & 12.20 & 12.35 & 13.44 & 66.17
\\
\arrayrulecolor{black!5}\hline
Base w/ Table \& Text & \multirow{2}{*}{40.22} & \multirow{2}{*}{53.47} & \multirow{2}{*}{54.18} & \multirow{2}{*}{55.57} & \multirow{2}{*}{81.18} &
\multirow{2}{*}{42.41} &
\multirow{2}{*}{54.06} &
\multirow{2}{*}{54.04} &
\multirow{2}{*}{\textbf{55.8}} & \multirow{2}{*}{81.63}
\\ \cite{zhang2023tablellama}

\\

\arrayrulecolor{black!5}\hline
Base w/ Table & \multirow{2}{*}{41.19} & \multirow{2}{*}{51.64} & \multirow{2}{*}{52.03} & \multirow{2}{*}{53.12} & \multirow{2}{*}{81.05} &
\multirow{2}{*}{37.34} &
\multirow{2}{*}{49.58} &
\multirow{2}{*}{49.80} &
\multirow{2}{*}{51.73} & 
\multirow{2}{*}{79.87}
\\ \& Summarized Text\setcounter{footnote}{1}\footnotemark 
\\

\arrayrulecolor{black}\hline
\rowcolor{blue!6}
Our Method w/o hopwise & 41.8 & 52.37 & 52.82 & 53.72 & 81.31 & 42.19 & 53.61 & 54.18 & 55.30 & 81.58
\\
\rowcolor{blue!6}
Our Method w/ hopwise & \textbf{44.2} & \textbf{55.82} & \textbf{55.28} & \textbf{56.90} & \textbf{83.98} & \textbf{44.30} & \textbf{54.08} &
\textbf{55.04} & 54.67 & \textbf{81.73}
\\

\arrayrulecolor{black}\hline
\rowcolor{gray!15}
\multicolumn{11}{c}{\centering \fontfamily{lmss}\selectfont{ \textit{Reader: Llama3-8B (zero-shot)}}} \\
\cline{1-11} 
Base & 2.0 & 7.07 & 6.91 & 7.07 & 59.05 & 0.64 & 7.00 & 6.77 & 8.72 & 59.71
\\
\arrayrulecolor{black!5}\hline
Base w/ Table \& Text &  
\multirow{2}{*}{28.6} &
\multirow{2}{*}{37.05} &
\multirow{2}{*}{37.22} &
\multirow{2}{*}{48.07} &
\multirow{2}{*}{74.01} &
\multirow{2}{*}{33.12} &
\multirow{2}{*}{43.43} &
\multirow{2}{*}{43.53} &
\multirow{2}{*}{45.12} &
\multirow{2}{*}{76.75}
\\ \cite{zhang2023tablellama}
\\

\arrayrulecolor{black!5}\hline
Base w/ Table & \multirow{2}{*}{30.33} & \multirow{2}{*}{39.42} & \multirow{2}{*}{39.60} & \multirow{2}{*}{41.30} & \multirow{2}{*}{75.06} &
\multirow{2}{*}{31.22} &
\multirow{2}{*}{41.72} &
\multirow{2}{*}{42.42} &
\multirow{2}{*}{42.88} &
\multirow{2}{*}{75.57}
\\ \& Summarized Text\setcounter{footnote}{1}\footnotemark 
\\
\arrayrulecolor{black}\hline
\rowcolor{blue!6}
Our Method w/o hopwise & 33.2  & 41.37 & 41.77 & 42.95 & 75.15 & 36.08 & 45.75 & 46.60 & 45.75 & 77.04
\\
\rowcolor{blue!6}

Our Method w/ hopwise 
& \textbf{37.0} & \textbf{46.43} & \textbf{46.56} & \textbf{48.78} & \textbf{77.55}  & \textbf{37.13} & \textbf{47.38} & \textbf{48.24} & \textbf{48.31} & \textbf{77.62}
\\
\arrayrulecolor{black}\hline
\end{tabularx}
\caption{\textbf{Table-Text QA Evaluation:} We analyze Exact Match (EM), F1-Score, Precision (P), Recall (R), and BERTScore-F1 (B) in (\%) to compare our method against baselines in a zero-shot setting using Llama3-8B, GPT-3.5, and GPT-4. The results consistently demonstrate significant improvements across datasets, metrics, and various language models. \textit{Base} (only reader LLM); \textit{w/ Table \& Text} (table and passages relevant to the question); \textit{w/ Table \& Summarized Text} (table with summarized supporting passages); \textit{w/o hopwise} (pruned information without considering hop-wise extraction).}
\label{tab:comparison}
\end{table*}

\renewcommand{\arraystretch}{1.0} 
\begin{table}[!ht]
    \centering
    \footnotesize
    \captionsetup{font=footnotesize}
    \rowcolors{1}{gray!15}{white}
    \begin{tabular}{p{5.4cm}p{0.7cm}p{0.7cm}} 
        \hline
        \textbf{\centering Method} & \textbf{\centering EM ($\uparrow$)} & \textbf{\centering F1 ($\uparrow$)} \\
        \hline
        \multicolumn{3}{c}{\textit{Hybrid-QA Fine-Tuning}} \\
        \hline
        HYBRIDER \cite{chen2020hybridqa} & 43.5 & 50.6 \\
        HYBRIDER-LARGE \cite{chen2020hybridqa} & 44.0 & 50.7 \\
        DocHopper \cite{sun2021iterative} & 47.7 & 55.0 \\
        MuGER\textsuperscript{2} \cite{wang2022muger2} & 57.1 & 67.3 \\
        \textbf{S\textsuperscript{3}HQA \cite{lei2023s3hqa} [SoTA]} & \textbf{68.4} & \textbf{75.3} \\
        \hline
        \multicolumn{3}{c}{\textit{w/o Fine-Tuning}} \\
        \hline
        Unsupervised-QG \cite{pan2021unsupervised} & 25.7 & 30.5 \\
                \hline
        GPT-4\textsuperscript{\textdagger}  \textit{w.} Retriever \cite{shi2024exploring} & 24.5 & 30.0 \\
                      \hline
        GPT-4\textsuperscript{\textdagger} + CoT \cite{wei2022chain} & 48.5 & 63.0 \\
        \hline
        HP\scalebox{0.8}{RO}P\scalebox{0.8}{RO}\textsuperscript{\textdagger}
        (\citet{shi2024exploring}, \textcolor{blue}{ACL 2024}) & 48.0 & 54.6 \\
        \hline       
        \textbf{ODYSSEY\textsuperscript{\textdagger} (Our Method)} & \textbf{51.5} & \textbf{66.0} \\
        \hline
    \end{tabular}
    \caption{Performance on Hybrid-QA validation set. \textsuperscript{\textdagger} stands for running on 200 sampled cases using gpt4-0613.}
    \label{tab:hybridqa_methods}
\end{table}

\begin{table}[!ht]
    \centering
    \footnotesize
    \captionsetup{font=footnotesize}
    \rowcolors{1}{gray!15}{white}
    \begin{tabular}{p{5.4cm}p{0.7cm}p{0.7cm}}
        \hline
        \textbf{\centering Method} & \textbf{\centering EM ($\uparrow$)} & \textbf{\centering F1 ($\uparrow$)} \\
        \hline
        \multicolumn{3}{c}{\textit{OTT-QA Fine-Tuning}} \\
        \hline
        BM25-HYBRIDER \cite{chen2020open} & 10.3 & 13.0 \\
        Fusion+Cross-Reader \cite{chen2020open} & 28.1 & 32.5 \\
        CARP \cite{zhong2022reasoning} & 33.2 & 38.6 \\
        CORE \cite{ma2022open} & 49.0 & 55.7 \\
        \textbf{COS \cite{ma2023chain} [SoTA]} & \textbf{56.9} & \textbf{63.2} \\
        \hline
        \multicolumn{3}{c}{\textit{w/o Fine-Tuning}} \\
                           \hline
        GPT-4\textsuperscript{\textdagger}  + CoT \cite{wei2022chain} & 61.0 & 72.3 \\
        \hline
    \textbf{ODYSSEY\textsuperscript{\textdagger} (Our Method)} & \textbf{62.02} & \textbf{73.02}
        \\ \hline
    \end{tabular}
    \caption{Performance on OTT-QA validation set. \textsuperscript{\textdagger} stands for running on 500 sampled cases using gpt4-1106-preview.}
    \label{tab:ottqa_methods}
\end{table}


We show evaluation on Hybrid-QA and OTT-QA datasets in Table \ref{tab:comparison}, \ref{tab:hybridqa_methods}, \ref{tab:ottqa_methods}.

Table \ref{tab:comparison} shows the performance of \name, our method with hop-wise extraction (denoted as `w/ hopwise' in the table). It consistently outperforms all baseline methods across Exact Match (EM), F1-score, Precision, Recall and BERTScore-F1 on the Hybrid-QA and OTT-QA datasets for all three LLMs. 
This improvement underscores the effectiveness of our approach in efficiently extracting crucial information while effectively filtering noise from the table-text in a question specific manner. 
It is worth noting that our model performs best in the zero-shot setting for closed-source LLMs like GPT-3.5 and GPT-4, and also shows strong performance with the comparatively smaller open-source model Llama3-8B for the task of Table-Text QA.

As mentioned in Section \ref{LlmReader}, if the LLM outputs \texttt{`None'} after 3-hops, it becomes necessary to pass the entire table and text as input to the LLM. These occurrences, where graph traversal is not possible or the LLM requires the full context, account for approximately 10\% and 8\% of the experiment data for Hybrid-QA and OTT-QA, respectively, across all models. 
However, it is important to note that, even without utilizing the complete experimental data, we achieve an EM score of 58\% on HybridQA and 62.0\% on OTTQA using GPT-4.

Table \ref{tab:hybridqa_methods} and \ref{tab:ottqa_methods} compare our zero-shot method with other approaches, including fine-tuning, few-shot methods \cite{pan2021unsupervised, shi2024exploring}, and CoT \cite{wei2022chain}. Our method outperforms the fine-tuned models on the OTT-QA dataset and achieves comparable results on the Hybrid-QA dataset.
Table \ref{tab:hybridqa_methods} presents results on the Hybrid-QA development set, which includes a randomly selected subset of 200 samples, as employed by HP\scalebox{0.8}{RO}P\scalebox{0.8}{RO} \cite{shi2024exploring} with a temperature setting of 0. For the final LLM call, we utilized GPT-4-0613 (consistent with \citet{shi2024exploring}), and GPT-3.5-turbo for Question Analysis. Notably, our method demonstrates a 7.3\% improvement in exact match (EM) and a 20.9\% increase in F1 score compared to   HP\scalebox{0.8}{RO}P\scalebox{0.8}{RO}.


\subsection{Efficient Query Context Handling}


In complex QA tasks, such as Hybrid tabular and textual QA with LLMs like GPT-3.5 and GPT-4, retrieving relevant information from the data corpus is crucial. This enables the LLMs to better connect the links between the structured and unstructured data for more accurate QA. Our method achieves this task by effectively filtering out noise, which not only increases QA accuracy (as shown in Table \ref{tab:comparison}) but also results in a reduced reader input context size compared to the original context (as shown in Table \ref{tab:total_token_count}). 
Specifically, the Question Analysis component of our method, discussed in Section \ref{QAnalysis} requires on average an input token size of 989 for Hybrid-QA and 850 for OTT-QA respectively. 
This amounts to total input token sizes to 4846 and 3595, respectively, representing a 32.65\% and 38.7\% reduction compared to the original context for Hybrid-QA and OTT-QA datasets respectively. Additionally, the output token size is smaller across all methods; our method produces around 70 tokens, while the baselines typically produce 4 tokens.

\begin{figure}[h]
    \includegraphics[width=\linewidth]{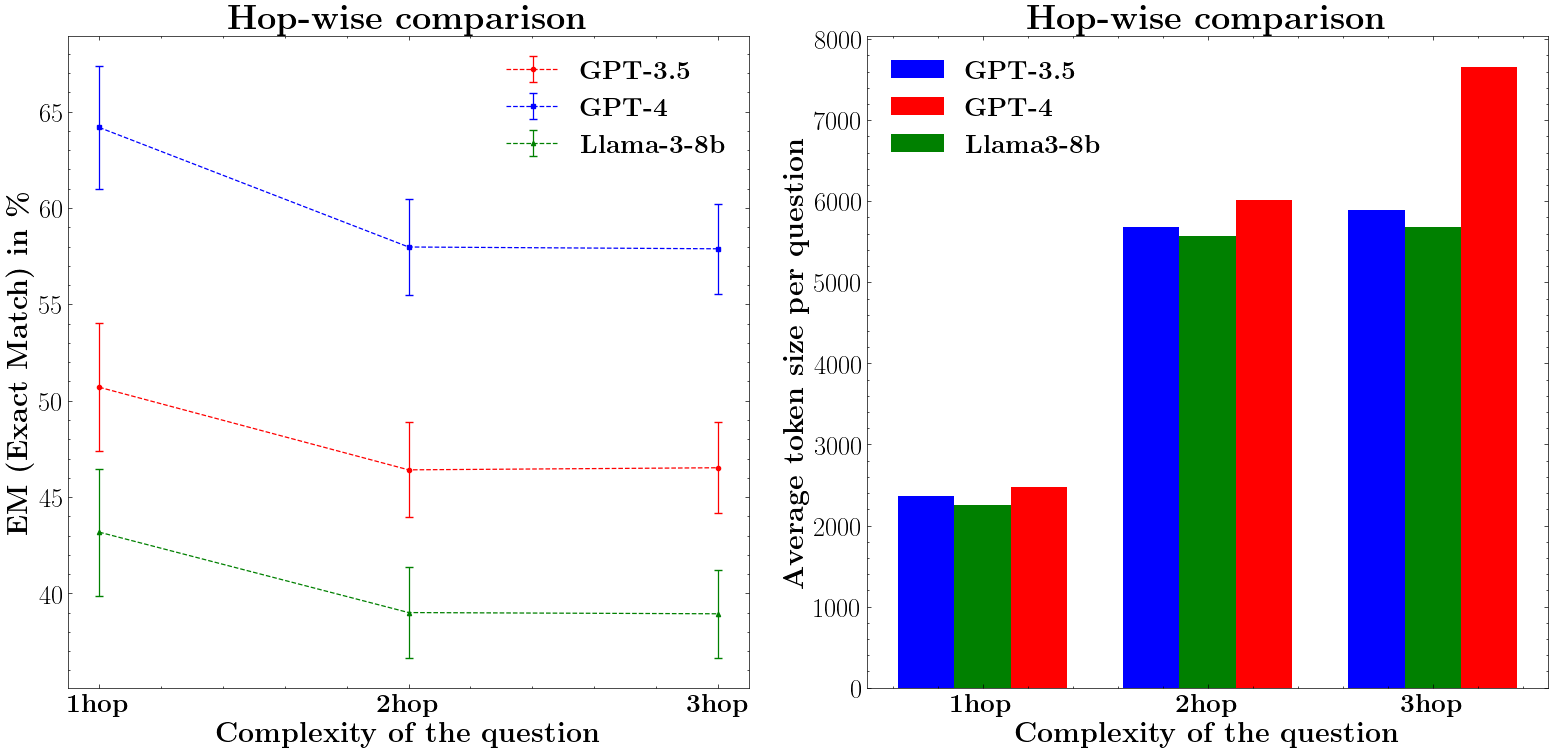}
    \caption{\textbf{Hopwise analysis:} For \name (our method w/ hopwise), we calculate the cumulative EM score (left-side in figure) and average token size (right-side in figure) utilized by each hop for Llama3-8B, GPT-3.5, and GPT-4 on Hybrid-QA. Bars in left-side of the figure denotes standard error -  $\sqrt{\frac{em(1-em)}{n}}$.}
  \label{fig:hopwise_plot}
\end{figure}

We analyze the questions answered by our method using a hop-wise criteria on the Hybrid-QA dataset (as shown in Figure \ref{fig:hopwise_plot}). Our method answers 144 questions with GPT-3.5 and 190 questions with GPT-4, achieving EM scores of 50.7\% and 64.8\% in 1-hop (see left-side of Figure \ref{fig:hopwise_plot}) with average token sizes of 1369 and 1479 (see right-side of Figure \ref{fig:hopwise_plot}), respectively. 
For both Hybrid-QA and OTT-QA, almost 90\% questions were answered using 1-hop and 2-hop connections.
This underscores the effectiveness of our Hybrid Graph-based method in conveying information to the LLM hop-wise. 


\renewcommand{\arraystretch}{1.1} 
\begin{table}[!ht]
    \centering
\begin{tabular}{lcc}
  \hline
 \textbf{Method}   & \textbf{Input Token} & \textbf{Input Token}  \\ 
  &
  \textbf{ Size ($\downarrow$)} & \textbf{Cost ($\downarrow$)}  
  \\ 
\cline{1-3}
\cline{1-3}
  \rowcolor{gray!3} 
\multicolumn{3}{c}{\centering \fontfamily{lmss}\selectfont{ \textit{Dataset: Hybrid-QA}}} \\
\cline{1-3} 
  Original Context & 7195 & \$71.95 \\
  Summarized & 3923 & \$39.23 \\
  Our Method &  3857 & \$38.57 \\
  \cline{1-3}
\cline{1-3}
  \rowcolor{gray!3} 
\multicolumn{3}{c}{\centering \fontfamily{lmss}\selectfont{ \textit{Dataset: OTT-QA}}} \\
\cline{1-3} 
  Original Context & 5866 & \$58.66 \\
  Summarized  & 3778 & \$37.78 \\
  Our Method   & 2745 &  \$27.45 \\
  \hline
\end{tabular}
\caption{\textbf{Reader Input Token Count and Cost}: 
We compare our method with baselines on average reader input token size and its pricing in dollars w.r.t. GPT-4 Turbo OpenAI pricing\protect\footnotemark[10] for 1000 samples.}
    \label{tab:total_token_count}

\end{table}




\subsection{Ablation Study}
\label{sec: ablation_study}







We present following ablation studies in this section: i) Hop-wise retrieval, and ii) Pruned Graph.
\\
\textbf{Hop-wise} We passed pruned information all at once instead of hop-wise retrieval. This means all retrieved information (up to 3-hops) was sent together to the large language model (LLM) for question answering. As hypothesized, the performance of our method remained nearly the same (see `Our Method w/o hopwise' in Table \ref{tab:comparison}), with a slight decrease observed for GPT-3.5 and Llama3-8b on the Hybrid-QA dataset. Consequently, the average input token size for readers increased by nearly 20\% for both datasets.
\\
\textbf{Pruned Graph} We passed the entire constructed Hybrid Graph to the LLM without pruning. 
Table \ref{tab:hybridqa_ottqa_comparison} in Appendix \ref{sec: comp_graph} shows results with complete hybrid graph in comparison to our method. The experiment conducted on the Hybrid-QA and OTT-QA dataset using GPT-4, resulted in a significant drop in performance compared to our method.
\footnotetext[10]{Pricing as of 15-June-2024 at \url{https://openai.com/api/pricing/}}

\subsection{Error Analysis}
\label{sec: error}

We performed a detailed error analysis on 100 random samples from the HybridQA development set using our method, \name, with GPT-4 in terms of EM score. Out of these, 41 answers were incorrect. The breakdown of our findings is as follows:

\begin{enumerate} \item \textbf{Expression Mismatch (17 cases):} Our method provided semantically correct answers expressed differently from the gold standard. For example, it might output "hosted by Regis Philbin," while the standard answer is simply "Regis Philbin." This discrepancy lowers the EM score despite the answers being fundamentally accurate.

\item \textbf{Semantic Module Errors (14 cases):}
\begin{itemize}[left=0pt, itemsep=0pt]
    \item \textbf{Entity-Matching (8 cases):} Errors occur when our model fails to match similar yet non-identical entities in the question and text. For instance, "1,301 acres" does not match "1,301," leading to potential misalignment with other numbers.
    
    \item \textbf{Entity-Extraction (2 cases):} These errors arise when our model struggles to extract complex or composite entities from the text.

    \item \textbf{Entity-Header Mapping (4 cases):} Due to complex questions, our method (using GPT) sometimes fails to identify all relevant table headers. We are addressing this by employing all headers when none are identified.
\end{itemize}

\item \textbf{LLM Errors (3 cases):} The Large Language Model occasionally fails to provide the correct answer, despite having the necessary context, especially in response to complex questions.

\item \textbf{Dataset Issues (7 cases):} These errors stem from ambiguous questions or anomalies in the dataset, rather than limitations of our method.

\end{enumerate}

\section{Conclusion}
In this paper, we introduce \name, a zero-shot fine-tuning-free approach for Table-Text QA. By leveraging a novel Hybrid Graph-based approach, \textit{Odyssey} effectively navigates the complexities of multi-hop reasoning across structured and unstructured data sources. Our method achieves significant improvements of 10\% and 4.5\% in Exact Match (EM) scores using GPT-3.5 on the Hybrid-QA and OTT-QA datasets, respectively, compared to the baseline, while reducing the input token size for the LLM reader by 45.5\% on Hybrid-QA and 53\% on OTT-QA, demonstrating its efficiency in representing relevant information concisely.
 We believe the insights gained from our method can pave the way for more advanced and efficient QA systems capable of navigating the ever-growing landscape of heterogeneous data sources.


\section*{Limitations}

\noindent The limitations of our work are as follows: 1) While our method achieves the highest accuracy in a zero-shot setting with Llama3-8B, GPT-3.5, and GPT-4, it incurs slightly more processing time than the zero-shot baseline due to additional LLM calls and Hybrid Graph traversal. 2) Our method's performance depends on current LLM capabilities, which may evolve over time. 3) We have only tested Tabular-Text data, whereas future work could explore various multi-modal datasets, including images and videos.

\section*{Acknowledgments}
We would like to thank Pranoy Panda and the other members of the AI Lab at Fujitsu Research India for their valuable feedback on the manuscript. Our sincere thanks also go to Prof. Pushpak Bhattacharyya, Lokesh N, and Nihar Ranjan Sahoo from IIT Bombay for their insightful comments, which helped improve the final draft.

\appendix

\section{Appendix}
\label{sec:appendix}

In this section, we provide additional results and details that we could not include in the main paper due to space constraints. In particular, this appendix contains the following:

\begin{itemize}[nosep]

    \item \hyperref[sec: additional_results]{Additional Results}
        \item \hyperref[sec: walkthrough]{\name Walkthrough with a Detailed Example}
    \item \hyperref[sec: prompts_our_method]{Prompts used for \name}
    \item \hyperref[sec:baseline_prompts]{Prompts used for baselines}

\end{itemize}




\subsection{Additional Results}
\label{sec: additional_results}

This section is divided into 2 parts: i) Input Complete Hybrid Graph, and ii) Results on Summarizing both the table and text. Additionally, it includes a table comparing the execution time per instance between our method and the baselines (see Table \ref{tab:time}).

\subsubsection{Input Complete Hybrid Graph}
\label{sec: comp_graph}
Table \ref{tab:hybridqa_ottqa_comparison} compares our method with the Complete Hybrid Graph approach, where the entire Hybrid Graph is passed directly without any traversal or reduction. This experiment evaluates whether GPT-like LLMs can infer relationships between graph nodes based solely on the question, without additional context. The results demonstrate that our method surpasses the Complete Hybrid Graph in both EM and F1 scores while using fewer input tokens. The Complete Hybrid Graph method failed to provide answers for over 40\% of the questions, and requiring larger token sizes due to full table-text passage inclusion. In contrast, our method is more accurate and efficient

\renewcommand{\arraystretch}{1.1} 
\begin{table*}[t]
\footnotesize
\captionsetup{font=footnotesize}
\centering
\begin{tabularx}{\textwidth}{l*{3}{>{\centering\arraybackslash}X}*{3}{>{\centering\arraybackslash}X}}
\cline{1-7}     
\multicolumn{1}{l}{\textbf{Datasets}} 
& \multicolumn{3}{c}{Hybrid-QA} 
& \multicolumn{3}{c}{OTT-QA} 
\\
\cmidrule(lr){2-4} \cmidrule(lr){5-7} 
\multicolumn{1}{l}{\textbf{ Methods}}
& \multicolumn{1}{p{0.4cm}}{\centering EM\\($\uparrow$)} 
& \multicolumn{1}{p{0.4cm}}{\centering F1\\($\uparrow$)} 
& \multicolumn{1}{p{0.8cm}}{\centering Input Token Size($\downarrow$)}
& \multicolumn{1}{p{0.4cm}}{\centering EM\\($\uparrow$)} 
& \multicolumn{1}{p{0.4cm}}{\centering F1\\($\uparrow$)} 
& \multicolumn{1}{p{0.8cm}}{\centering Input Token Size($\downarrow$)}
\\ 
\cline{1-7}
\cline{1-7}
\rowcolor{gray!15} 
\multicolumn{7}{c}{\centering \fontfamily{lmss}\selectfont{ \textit{Reader: gpt-4-1106-preview (zero-shot)}}} \\
\cline{1-7}

Complete Hybrid Graph & 54.0 & 65.30 & 5449 & 48.0 & 64.12 & 5763 
\\
Our Method & \textbf{58.0} & \textbf{71.80} & \textbf{3857} & \textbf{59.0} & \textbf{69.84} & \textbf{2745} 
\\
\arrayrulecolor{black}\hline
\end{tabularx}
\caption{\textbf{Comparison of EM, F1, and Avg Token Size across Hybrid-QA and OTT-QA datasets with GPT-4 model:} The table compares the performance between the Complete Hybrid Graph as Input and Our Method.}
\label{tab:hybridqa_ottqa_comparison}
\end{table*}

\subsubsection{Summarising both the table and text}
\label{sec: summarizing}

 For this baseline, we adopted the method proposed by LangChain\footnotemark[\value{footnote}], which involves summarizing tables and text and pass them as input to the LLM for QA. However, converting structured tables into summarized text led to a loss of information, resulting in lower scores (see Table \ref{tab:langchain_summarized}). Although summarizing the context reduces token size, it often fails to effectively filter out noise and may erase relevant information. We address this limitation in our method by pruning relevant information.

\renewcommand{\arraystretch}{1.1} 
\begin{table}[!ht]
    \centering
    \footnotesize
    \captionsetup{font=footnotesize}
    \begin{tabular}{lccc}
    \hline
        \textbf{Methods} & \textbf{EM ($\uparrow$)} & \textbf{F1 ($\uparrow$)} & \textbf{B ($\uparrow$)} \\
    \hline
        GPT-4-1106-preview & 31.92 & 43.19 & 74.64 \\
        GPT-3.5-turbo-1106  & 30.1 & 41.6 & 75.13 \\
        Llama3-8B & 21.11 & 29.02 & 70.61 \\
    \hline
    \end{tabular}
    \caption{Results for context w/ summarized table and summarized passages for the Hybrid-QA dataset.}
    \label{tab:langchain_summarized}
\end{table}

\renewcommand{\arraystretch}{1.1} 
\begin{table}[!ht]
    \centering
    \footnotesize
    \captionsetup{font=footnotesize}
    \begin{tabular}{lccc}
    \hline
        \textbf{Methods} & \textbf{Time (in seconds)} \\
    \hline
        Base & 1 \\
        Base w/ Table \& Text  & 1.5 \\
        Base w/ Table and Summarized Text & 7 \\
           Our Method w/o hopwise &  3.6\\
           Our Method w/ hopwise & 5 \\
    \hline
    \end{tabular}
    \caption{Per-instance execution time of each method presented in the main paper (see Table \ref{tab:comparison}).}
    \label{tab:time}
\end{table}

\subsection{\name Walkthrough with a Detailed Example}
\label{sec: walkthrough}

 \texttt{\noindent
In this section, we will understand \name with an illustrated example.
\\
Below, we describe the context of the chosen example, including the question, table, and passages.
\\
\textbf{Question:} The driver who finished in position 4 in the 2004 United States Grand Prix was of what nationality ? 
\\
\textbf{Table Headers:} [Pos, Driver, Constructor, Time, Gap]
\\
The table contains 20 rows, with some cells linked to passages. We will list some of these passages below.}
\\
\texttt{
\textbf{Passages:}}
\begin{enumerate}[nosep]

    \item \texttt{ Michael\_Schumacher :  Michael Schumacher ( born 3 January 1969 ) is a retired German racing driver who raced in Formula One for Jordan Grand Prix , Benetton and Ferrari , where he spent most of his career , as well as for Mercedes upon his return to the sport . Widely regarded as one of the greatest Formula One drivers ever , and regarded by some as the greatest of all time , Schumacher is the only driver in history to win seven Formula One World Championships , five of which he won consecutively ....}

    \item \texttt{ Jenson\_Button: Jenson Alexander Lyons Button MBE ( born 19 January 1980 ) is a \textbf{British racing driver} and former Formula One driver . He won the 2009 Formula One World Championship , driving for Brawn GP . He currently competes in the Japanese Super GT Series driving a Honda NSX-GT for Team Kunimitsu , in which he won the title in 2018 . Button began karting at the age of eight and achieved early success ....}

\end{enumerate}

\noindent \textbf{\texttt{Question Analysis}}
\\
\texttt{
In this phase, we extract relevant entities from the question, identify pertinent headers from all table headers in relation to the question, and then map the extracted entities to the identified headers.}
\\
\texttt{
\textbf{Entity Extraction from the Question:} [`driver', `position 4', `2004 United States Grand Prix', `nationality']
\\
\textbf{Relevant Headers:} [`Pos', `Driver']
\\
\textbf{Entity-Header Mapping:} \{
  `driver': [`Driver'], 
  `position 4': [`Pos'], 
  `2004 United States Grand Prix': [`Others'], 
  `nationality': [`Others']
\}}
\\
\\
\noindent \textbf{\texttt{Hybrid Graph Construction}}
\\
\texttt{
After completing the question analysis phase, our next step is to jointly connect the table and passages for efficient retrieval. To achieve this, we construct the Hybrid Graph.}
\\
\texttt{For Hybrid Graph construction, there are 2 processes:}
\begin{itemize}[nosep]
    \item \texttt{Table Retrieval: The relevant table headers are selected as sub-table, which becomes part of the Hybrid Graph.}

    \item \texttt{Passage Retrieval: Entities are extracted from all the passages, and an Entity-Document graph is created, linking each extracted entity to its corresponding passage or document.}
    \\
    \texttt{Example of entities extracted: [`Jenson Alexander Lyons Button MBE', `19 January 1980', `British', `Formula One', `2009', `Formula One World Championship', `Brawn GP', ..]}
\end{itemize}
 
\texttt{ \noindent After retrieving the sub-table and creating the Entity-Document graph, we connect the entities from the passages to their corresponding table cells, forming the Hybrid Graph.}
\\
\\
\texttt{
\noindent \textbf{Hybrid Graph Traversal}
\\
With the entities extracted from the question, we traverse the Hybrid Graph up to 3 hops and store the results in a hop-wise dictionary. A sample of the dictionary is shown below for our example:
\\
\{"1-hop": ["4; Pos", "Driver", "Jenson Button"], "2-hop": ["Jenson Button; Driver", "entity", "British"], "3-hop":  ["British", "Driver", "Juan Pablo Montoya"]\}
\\
In the dictionary, within each tuple, the word following `;' represents the header for the cell, while we track the entities originating from the table and those extracted from the passages.
}
\\
\\
\texttt{
\noindent \textbf{LLM Reader}}
\\
\texttt{After obtaining the traversed graph, we pass it to the LLM hop-wise. Before providing it as input to the LLM, we preprocess our graph paths to form a sub-table, including the corresponding linked passages.}
\par
\texttt{For this example, we pass the question along with the 1st hop of the dictionary. After preprocessing, we obtain the sub-table and the passages attached to the cells in the sub-table, which are then passed to the reader LLM for QA.
The LLM identifies that `Jenson Button' is the driver who achieved the position, and from the corresponding passage, it retrieves his nationality as `British,' providing this as the output.
}

\onecolumn
\subsection{Prompts used for \name}
\label{sec: prompts_our_method}

This section details the prompts used for our method, \name, and is divided into: i) Question Analysis and ii) Final LLM Reader call.

\subsubsection{Question Analysis}
\label{sec: QAnalysis_prompt}

We start with Question Analysis, which uses a 1-shot example and follows three steps:
\begin{enumerate}[nosep, label=\roman*.]
    \item Entity Extraction - Extract entities from the question, using the table name and headers as additional input for efficiency.
    \item Relevant Table Headers - Extract useful headers from the table for answering the question, using the extracted entities as additional input.
    \item Entity-Header Mapping - Map the extracted question entities to the fetched table headers.
\end{enumerate}

\begin{tcolorbox}[
    enhanced,
    breakable,
     width=\textwidth, 
    title=Entity Extraction from the Question,
    fonttitle=\bfseries\large,
    colframe=blue!75!black, 
    colback=white!10!white, 
    coltitle=white, 
    colbacktitle=blue!85!black, 
    boxrule=0.2mm,
    sharp corners,
    shadow={1mm}{-1mm}{0mm}{black!50!white}, 
    attach boxed title to top left={yshift=-3mm, xshift=3mm},
    boxed title style={sharp corners, size=small}
]
\textbf{Agent Introduction:} You are an agent who is going to be assisting me in a question answering task. For this task, I need to first identify the named entities in the question.

\vspace{2mm} 
\textbf{Task:} Identify the named entities in the provided question. These entities will serve as key elements for extracting pertinent information from the available sources, which include table name and its headers.

\vspace{2mm} 
\textbf{Output format:} 
\[
\begin{array}{l}
\text{Entities: [`<entity1>', `<entity2>', .....]}\\
\end{array}
\]

\vspace{2mm} 
\textbf{Use the below example to better understand the task}
\vspace{1mm} 
\\
\textbf{Input:} 
\\
Question: What was the nickname of the gold medal winner in the men's heavyweight Greco-Roman wrestling event of the 1932 Summer Olympics? 
\\
Table Name: Sweden at the 1932 Summer Olympics 
\\
Table Headers: ["Medal", "Name", "Sport", "Event"]

\vspace{3mm} 
\textbf{Output:} 
\[
\begin{array}{l}
\text{Entities: [`nickname', `medal', `gold', `men's heavyweight',} \\
\text{`Greco-Roman Wrestling event', `1932 Summer Olympics']}\\
\end{array}
\]
\vspace{3mm} 
\textbf{Input:}\\
Question: \{question\}\\
Table Name: \{table\_id\}\\
Table Headers: \{table\_headers\}

\vspace{1mm} 
\textbf{Output:}
\end{tcolorbox}

\newpage

\begin{tcolorbox}[
    enhanced,
    breakable,
    width=\textwidth, 
    title=Relevant Header Extraction,
    fonttitle=\bfseries\large,
    colframe=blue!75!black, 
    colback=white!10!white, 
    coltitle=white, 
    colbacktitle=blue!85!black, 
    boxrule=0.2mm,
    sharp corners,
    shadow={1mm}{-1mm}{0mm}{black!50!white}, 
    attach boxed title to top left={yshift=-3mm, xshift=3mm},
    boxed title style={sharp corners, size=small}
]
\textbf{Agent Introduction:} You are an agent who is going to be assisting me in a question answering task. I have a table as a source of information. I have already extracted the relevant entities from the question. For this task, I need to first identify the column headers that are relevant in the question.

\vspace{2mm} 
\textbf{Task:} 
Identify the relevant column headers from the provided list, based on the extracted entities from the question.  I will also provide the extracted entities from the question and name of the table.

\vspace{2mm} 
\textbf{Output format:} 
\[
\begin{array}{l}
\text{Relevant headers: [`<header-1>', `<header-2>', ....]}\\
\end{array}
\]
\vspace{1mm}
\\
\textbf{Use the below example to better understand the task}
\vspace{1mm} 
\\
\textbf{Input:} 
\\
Question: What was the nickname of the gold medal winner in the men's heavyweight Greco-Roman wrestling event of the 1932 Summer Olympics? 
\\
Table Name: Sweden at the 1932 Summer Olympics 
\\
Table Headers: ["Medal", "Name", "Sport", "Event"] 
\\
Entities extracted from question: ["gold medal", "men's heavyweight", "Greco-Roman Wrestling", "1932 Summer Olympics"]

\vspace{3mm} 
\textbf{Output:} 
\[
\begin{array}{l}
\text{Relevant headers: ["Medal", "Name", "Sport", "Event"]}\\
\end{array}
\]
\vspace{3mm} 
\textbf{Input:} 
\\
Question: \{question\}
\\
Table Name: \{table\_id\}
\\
Table Headers: \{table\_headers\}
\\
Entities extracted from question: \{entities\}

\vspace{2mm} 
\textbf{Output:}
\end{tcolorbox}


\begin{tcolorbox}[
    enhanced,
    breakable,
    width=\textwidth, 
    title=Entity to Header Mapping,
    fonttitle=\bfseries\large,
    colframe=blue!75!black, 
    colback=white!10!white, 
    coltitle=white, 
    colbacktitle=blue!85!black, 
    boxrule=0.2mm,
    sharp corners,
    shadow={1mm}{-1mm}{0mm}{black!50!white}, 
    attach boxed title to top left={yshift=-3mm, xshift=3mm},
    boxed title style={sharp corners, size=small}
]
\textbf{Agent Introduction:} You are an agent who is going to be assisting me in a question answering task. I have a table as a source of information. I have already extracted relevant entities from the question and relevant column headers from the table.

\vspace{2mm} 
\textbf{Task:} 
Map the entities extracted from the question with the relevant headers and the table name. 

\vspace{2mm} 
\textbf{Output format:} 
\[
\begin{array}{l}
\text{"<entity1>": ["<mapping1>", "<mapping2>"],} \\
\text{"<entity2>": ["<mapping1>"]} \\
\end{array}
\]
For each entity extracted from the question, there should be a corresponding <mapping> to an item in the `Relevant headers' column. If none of the headers match the entity, the mapping should be labeled as "Others".

\vspace{1mm}
\textbf{Use the below example to better understand the task}
\vspace{1mm} 
\\
\textbf{Input:} 
\\
Question: What was the nickname of the gold medal winner in the men's heavyweight Greco-Roman wrestling event of the 1932 Summer Olympics? 
\\
Table Name: Sweden at the 1932 Summer Olympics 
\\
Entities extracted from question: ["gold medal", "men's heavyweight", "Greco-Roman Wrestling", "1932 Summer Olympics"]
\\
Relevant headers: ["Medal", "Name", "Sport", "Event"]

\vspace{3mm} 
\textbf{Output:} 
\[
\begin{array}{l}
\text{"gold medal": ["Medal"],} \\
\text{"men's heavyweight": ["Event"],} \\
\text{"Greco-Roman Wrestling": ["Sport"],} \\
\text{"1932 Summer Olympics": ["Others"]} \\
\end{array}
\]

\vspace{3mm} 
\textbf{Input:} 
\\
Question: \{question\}
\\
Table Name: \{table\_id\}
\\
Entities extracted from question: \{entities\}
\\
Relevant Headers: \{relevant\_headers\}

\vspace{2mm} 
\textbf{Output:}
\end{tcolorbox}

\subsubsection{LLM Reader} \label{sec:prompt_reader}
This is the final LLM call after Question Analysis for QA, operating in a zero-shot setting.

\begin{tcolorbox}[
    enhanced,
    breakable,
    width=\textwidth, 
    title = Hybrid Question Answering for \name,
    fonttitle=\bfseries\large,
    colframe=blue!75!black, 
    colback=white!10!white, 
    coltitle=white, 
    colbacktitle=blue!85!black, 
    boxrule=0.2mm,
    sharp corners,
    shadow={1mm}{-1mm}{0mm}{black!50!white}, 
    attach boxed title to top left={yshift=-3mm, xshift=3mm},
    boxed title style={sharp corners, size=small}
]
\textbf{Agent Introduction:} 
Hello! I'm your Hybrid-QA expert agent, here to assist you in answering complex questions by leveraging both table data and passage information. Let's combine these sources to generate accurate and comprehensive answers!

\vspace{2mm} 
\textbf{Task:} 
Your task involves a central question that requires information from both a table and passages.

\vspace{2mm} 
Here's the context you'll need:

\vspace{1mm}
\textbf{Table Data:} 
\{table\_data\}

\vspace{1mm}
\textbf{Passages:} 
\{passages\}

\vspace{1mm}
\textbf{Question:} 
\{question\}

\vspace{2mm} 
    \textbf{Final Answer:}
        Provide the final answer in the format below. If the answer cannot be answered with the given context, provide None.

\vspace{1mm}
    \textbf{Final Answer Format:}
    \[
\begin{array}{l}
        \text{Final Answer: <your answer>}
\end{array}
    \]
\vspace{2mm} 
    If the final answer is "None", provide the names of passages that are relevant to the above questions. If no passages are relevant give `[]' as Relevant Passages.
    
\vspace{1mm}
    \textbf{Relevant Passages Format:}
    \[
\begin{array}{l}
       \text{ Relevant Passages: [`<name-of-passage1>', `<name-of-passage2>', ......]}
\end{array}
    \]
        
\end{tcolorbox}

\newpage

\subsection{Prompts used for baselines}
\label{sec:baseline_prompts}

We compare our method with two baselines using context for QA: i) Summarized Context and ii) Complete Context. For table summarization, we pass the entire table to the LLM for summarization. For text, each passage is passed individually. The final LLM reader prompt is the same for both methods.

\subsubsection{Table and Text Summarization}


\begin{tcolorbox}[
    enhanced,
    breakable,
    width=\textwidth, 
    title=Table Summarization Task,
    fonttitle=\bfseries\large,
    colframe=blue!75!black, 
    colback=white!10!white, 
    coltitle=white, 
    colbacktitle=blue!85!black, 
    boxrule=0.2mm,
    sharp corners,
    shadow={1mm}{-1mm}{0mm}{black!50!white}, 
    attach boxed title to top left={yshift=-3mm, xshift=3mm},
    boxed title style={sharp corners, size=small}
]
\textbf{Agent Introduction:} 
You are an assistant tasked with summarizing tables.

\vspace{2mm} 
\textbf{Task:} 
I have a table that I need help summarizing. The table contains the following columns and data: \{table\}
\\
Provide a concise summary of the table without removing any numbers, entities or important information. Try to retain all the important information. 

\vspace{2mm} 
\textbf{Response:} 

\end{tcolorbox}


\begin{tcolorbox}[
    enhanced,
    breakable,
    width=\textwidth, 
    title=Passage Summarization Task,
    fonttitle=\bfseries\large,
    colframe=blue!75!black, 
    colback=white!10!white, 
    coltitle=white, 
    colbacktitle=blue!85!black, 
    boxrule=0.2mm,
    sharp corners,
    shadow={1mm}{-1mm}{0mm}{black!50!white}, 
    attach boxed title to top left={yshift=-3mm, xshift=3mm},
    boxed title style={sharp corners, size=small}
]
\textbf{Agent Introduction:} 
You are an assistant tasked with summarizing passages.

\vspace{2mm} 
\textbf{Task:} 
I have a passage that I need help summarizing. The passage is as follows: \{text\}
\\
Provide a concise summary of the passage without removing any numbers, entities or important information. Try to retain all the important information. 

\vspace{2mm} 
\textbf{Response:} 

\end{tcolorbox}

\subsubsection{LLM Reader}

This is the LLM call for baselines operating in a zero-shot setting.

\begin{tcolorbox}[
    enhanced,
    breakable,
    width=\textwidth, 
    title = Hybrid Question Answering,
    fonttitle=\bfseries\large,
    colframe=blue!75!black, 
    colback=white!10!white, 
    coltitle=white, 
    colbacktitle=blue!85!black, 
    boxrule=0.2mm,
    sharp corners,
    shadow={1mm}{-1mm}{0mm}{black!50!white}, 
    attach boxed title to top left={yshift=-3mm, xshift=3mm},
    boxed title style={sharp corners, size=small}
]
\textbf{Agent Introduction:} 
Hello! I'm your Hybrid-QA expert agent, here to assist you in answering complex questions by leveraging both table data and passage information. Let's combine these sources to generate accurate and comprehensive answers!

\vspace{2mm} 
\textbf{Task:} 
Your task involves a central question that requires information from both a table and passages.

\vspace{2mm} 
Here's the context you'll need:

\vspace{1mm}
\textbf{Table Data:} 
\{table\_data\}

\vspace{1mm}
\textbf{Passages:} 
\{passages\}

\vspace{1mm}
\textbf{Question:} 
\{question\}

\vspace{2mm} 
    \textbf{Final Answer:} <your answer>

\end{tcolorbox}

\end{document}